%% file: main.tex
\documentclass[10pt,twocolumn,letterpaper]{article}

\usepackage[pagenumbers]{cvpr} 

\usepackage{graphicx}
\usepackage{amsmath}
\usepackage{amssymb}
\usepackage{booktabs}
\usepackage{multirow}

\usepackage[pagebackref,breaklinks,colorlinks]{hyperref}

\usepackage[capitalize]{cleveref}
\crefname{section}{Sec.}{Secs.}
\Crefname{section}{Section}{Sections}
\Crefname{table}{Table}{Tables}
\crefname{table}{Tab.}{Tabs.}

\newcommand{\tracker}{Immortal Tracker}
\newcommand{\trackerspace}{Immortal Tracker }

\def\cvprPaperID{3360} 
\def\confName{CVPR}
\def\confYear{2022}

\input{preamble}
\begin{document}
\input{0_metadata}
\maketitle
\input{0_abstract}
\input{1_introduction}

\input{2_related}

\input{3_method}

\input{4_results}
\input{5_conclusions}

{
    \small
    \bibliographystyle{ieee_fullname}
    \bibliography{macros,main}
}



\end{document}

%% file: preamble.tex
\pdfoutput=1
\usepackage{overpic}
\usepackage{enumitem} 
\usepackage{overpic} 
\usepackage{color}

\definecolor{turquoise}{cmyk}{0.65,0,0.1,0.3}
\definecolor{purple}{rgb}{0.65,0,0.65}
\definecolor{dark_green}{rgb}{0, 0.5, 0}
\definecolor{orange}{rgb}{0.8, 0.6, 0.2}
\definecolor{red}{rgb}{0.8, 0.2, 0.2}
\definecolor{darkred}{rgb}{0.6, 0.1, 0.05}
\definecolor{blueish}{rgb}{0.0, 0.3, .6}
\definecolor{light_gray}{rgb}{0.7, 0.7, .7}
\definecolor{pink}{rgb}{1, 0, 1}
\definecolor{greyblue}{rgb}{0.25, 0.25, 1}

\usepackage{blindtext}

\renewcommand{\paragraph}[1]{\vspace{1em}\noindent\textbf{#1}.}
\usepackage{algorithm}
\usepackage{algpseudocode}
\usepackage{makecell}
\usepackage{lipsum}

%% file: 0_metadata.tex
\title{\tracker: Tracklet Never Dies}

\author{
\text{Qitai Wang}$^{1,3}$ 
\hspace{3mm} \text{Yuntao Chen}$^{2}$ 
\hspace{3mm} \text{Ziqi Pang}$^{4}$
\hspace{3mm} \text{Naiyan Wang}$^2$ \hspace{3mm} 
\text{Zhaoxiang Zhang}$^{1,3}$\\
$^{1 }$ \text{ University of Chinese Academy of Sciences} \hspace{5mm} $^{2 }$\text{ Tusimple}\\ 
$^{3 }$ \text{ Institute of Automation, Chinese Academy of Sciences (CASIA)}\\
$^{4 }$ \text{ UIUC}\\
\tt\small{ $\{$wangqitai2020, zhaoxiang.zhang$\}$@ia.ac.cn} \\
\tt\small{$\{$chenyuntao08, winsty$\}$@gmail.com ziqip2@illinois.edu }
}


%% file: 0_abstract.tex
\begin{abstract}


Previous online 3D Multi-Object Tracking(3DMOT) methods terminate a tracklet when it is not associated with new detections for a few frames. 
But if an object just goes dark, like being temporarily occluded by other objects or simply getting out of FOV, terminating a tracklet prematurely will result in an identity switch.
We reveal that \textbf{premature tracklet termination} is the main cause of identity switches in modern 3DMOT systems.
To address this, we propose \tracker, a simple tracking system that utilizes trajectory prediction to maintain tracklets for objects gone dark. 
We employ a simple Kalman filter for trajectory prediction and preserve the tracklet by prediction when the target is not visible.
With this method, we can avoid 96\% vehicle identity switches resulting from premature tracklet termination.
Without any learned parameters, our method achieves a mismatch ratio at the 0.0001 level and competitive MOTA for the vehicle class on the Waymo Open Dataset test set.
Our mismatch ratio is tens of times lower than any previously published method. 
Similar results are reported on nuScenes. 
We believe the proposed \trackerspace can offer a simple yet powerful solution for pushing the limit of 3DMOT.
Our code is available at \href{https://github.com/ImmortalTracker/ImmortalTracker}{https://github.com/ImmortalTracker/ImmortalTracker}.

\end{abstract}


%% file: 1_introduction.tex
\section{Introduction}
\label{sec:intro}
%


Most state-of-the-art methods in 3D Multiple Object Tracking(3DMOT) follow the tracking-by-detection paradigm. 
They first obtain bounding boxes of objects in each frame with 3D detectors and then associate the detected objects between frames as tracklets. 
In the online tracking process, the same object is associated sequentially across frames. 

\input{introfig}

Ideally, a tracklet should be maintained until its target leaves the scene. 
Previous works \cite{Weng20203DMT,weng2020gnn3dmot,sun2020scalability,chiu2020probabilistic,kim2020eager,yin2021center,benbarka2021score,zaech2021learnable} in 3DMOT usually adopt a simple delay-based mechanism for tracklet life management. 
They maintain a tracklet that loses its target for a few frames.
If the tracklet fails to be associated with any detections during this period, it will be terminated and no longer participate in the detection-tracklet association in the future frames. 
This mechanism assumes that an object not detected for several frames is considered to have left the scene. 
However, objects go dark for different reasons.
When an object is occluded or temporarily being out of FOV, it might be missing for frames as well and its corresponding tracklet will be terminated prematurely. 
In such a case if the object is detected again in the future, a new tracklet will be initialized, causing an identity switch.
As shown in Table~\ref{tab:idswstatic}, 99.9\% vehicle identity switches reported by the CenterPoint tracker on the validation set of Waymo Open Dataset (WOD) are the results of premature tracklet termination.

To overcome premature tracklet termination, we present \tracker, a simple tracking framework based on trajectory prediction for invisible objects. 
Instead of terminating an unassociated tracklet, we maintain it with its predicted trajectory. 
Therefore when a temporarily invisible object reappears on the predicted trajectory, it can be assigned to its original tracklet. 
We use a vanilla 3D Kalman filter(3DKF)\cite{kalman1960new} for trajectory prediction to highlight the simplicity and effectiveness of our predict-to-track paradigm. 
As shown in Figure~\ref{fig:introfig}, our simple 3DKF works surprisingly well for trajectory prediction on WOD.

Experiments show that our \trackerspace can significantly reduce identity switches when applied on the top of a wide range of off-the-shelf 3D detectors. 
Built upon the detection results from CenterPoint\cite{yin2021center}, we establish new SOTA results on 3DMOT benchmarks like Waymo Open Dataset and nuScenes. 
Our method achieves 60.6 level 2 MOTA for the vehicle class on Waymo Open Dataset.
We also bring the mismatch ratio down to the 0.0001 level, tens of times lower than previous SOTA methods. 
On nuScenes, our method achieves a 66.1 AMOTA, outperforming all previous published LiDAR-based methods. 

\input{idsw_static}

%% file: introfig.tex
\begin{figure}[t]
\begin{center}
\includegraphics[width=\linewidth,trim=100 150 100 100,clip]{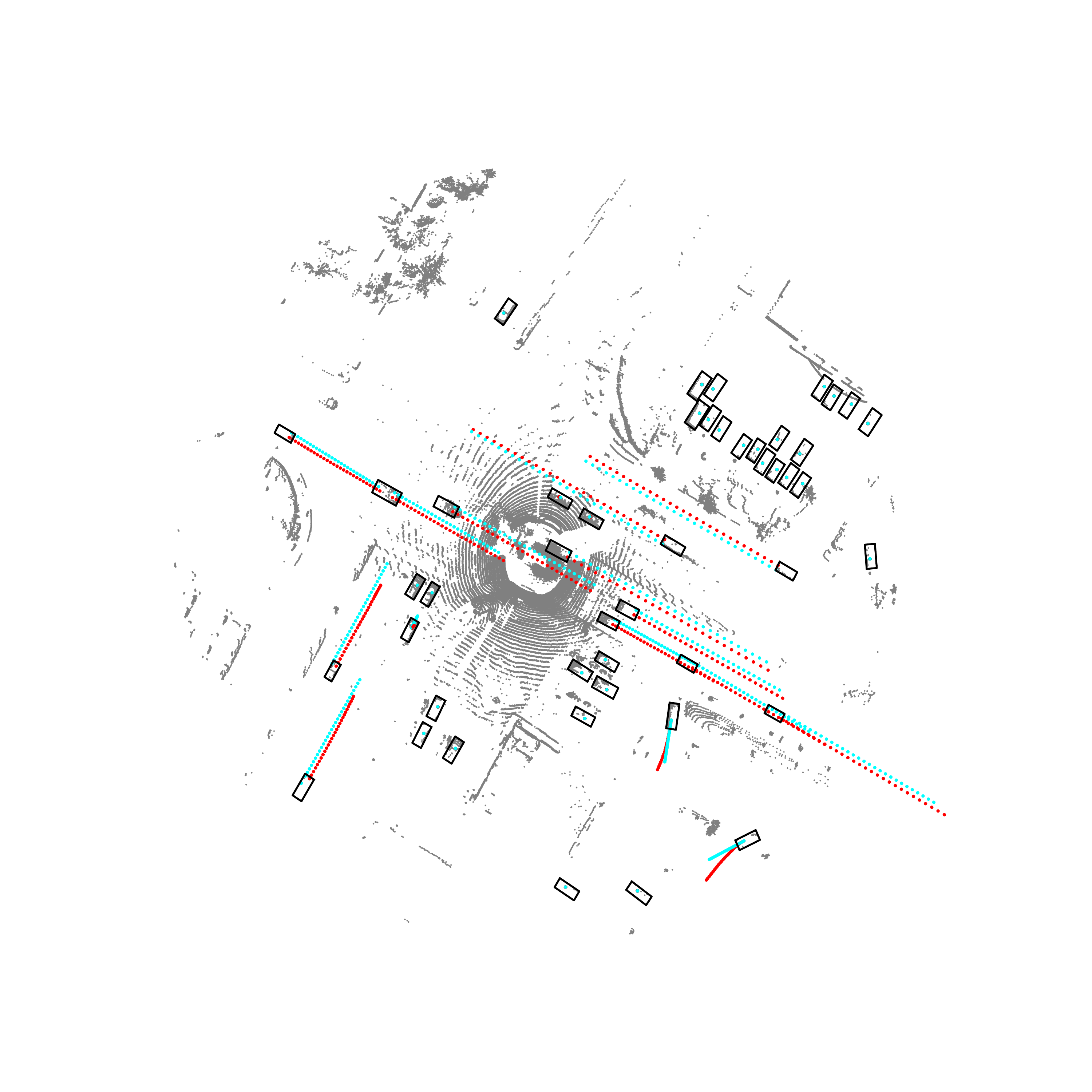}
\end{center}
\caption{
Bird’s Eye View(BEV) tracking visualization of vehicles with their predicted trajectories. For each tracked vehicle, we plot their bounding boxes in the initial frame and predicted(marked in cyan) or ground-truth(marked in red) center locations for the next 30 frames. Small displacements are added to the overlapped trajectories for better visualization.  
}
\label{fig:introfig}
\end{figure}

%% file: idsw_static.tex
\begin{table}\normalsize
\centering
\resizebox{\linewidth}{!}{ 
\begin{tabular}{@{}lcccccc@{}}
\toprule

  Method & IDS & \makecell[c]{Early\\ Termination} & \makecell[c]{Wrong\\ Assosication}\\
\midrule
CenterPoint & 2891 & 2890 & 1\\
\tracker(Ours)        & 114 & 113 &  1\\
\bottomrule
\end{tabular}
} 
\caption{
Statics on the vehicle identity switch cases reported by CenterPoint and \trackerspace on the validation set of WOD. We divide the identity switches into Early Termination(the ground truth trajectory of an object breaks into two tracklet) and Wrong Association(two tracklets belonging to different objects are confused during the tracking process). 
}
\label{tab:idswstatic}
\end{table}

%% file: 2_related.tex
\section{Related works}
\label{sec:related}

{\bf 3D MOT.}
Following the tracking-by-detection paradigm, previous works in 3D multi-object tracking usually solve the tracking problem in the form of a bipartite graph matching process on top of off-the-shelf detectors.
Inspired by early works in 2D MOT\cite{bewley2016simple,wojke2017simple,dicle2013way}, various methods focus on strengthening the association between detections and tracklets by modeling their motions or appearance, or the combination of these both.
\cite{patil2019h3d} defines the state of Kalman Filter in 2D plane. 
AB3DMOT\cite{Weng20203DMT} presents a baseline method through combining 3D Kalman filter and the Hungarian algorithm\cite{kuhn1955hungarian} on top of PointRCNN detector\cite{shi2019pointrcnn}. 
AB3DMOT employs 3D Intersection of Union  (3D IoU) as its association metric, while Chiu et al\cite{chiu2021probabilistic} used Mahalabnobis distance\cite{mahalanobis1936generalized} instead. SimpleTrack\cite{pang2021simpletrack} generalized GIoU\cite{Rezatofighi_2019_CVPR} to 3D for association. 
CenterPoint\cite{yin2021center} learns to predict two-dimensional velocity of detected box following CenterTrack\cite{zhou2020tracking} and perform simple point-distance matching. 

To further reduce confusion during the association process, GNN3DMOT\cite{weng2020gnn3dmot} jointly extracts appearance and motion features with a Graph Neural Network to introduce feature interactions between objects. 
\cite{chiu2021probabilistic} proposes a probabilistic multi-modal system that contains trainable modules for 2D-3D appearance feature fusion,  distance combination, and trajectory initialization. 
\cite{kim2020eager} combines 2D and 3D object evidence obtained from 2D and 3D detectors.  
Authors of \cite{zaech2021learnable} merge predictive models and objects detection features in a unified graph representation.
In this paper, we use simple 3D IoU/GIoU metrics for data association. 

Following early experience in 2D MOT\cite{bewley2016simple,wojke2017simple,he2021learnable}, previous works\cite{Weng20203DMT,weng2020gnn3dmot,chiu2020probabilistic,kim2020eager,yin2021center,zaech2021learnable} in 3D MOT usually adopt count-based method for tracklet life-cycle management. For each frame, new tracklets are initialized with detections that are not associated with existing tracklets. Tracklets that lost their targets for serveral(typically less than 5) frames will be terminated.
\cite{benbarka2021score,sun2020scalability} proposed to initialize and terminate tracks depending on their confidence score estimated from the confidence of their associated detections. However, they will still permanently terminate the tracklets that fail to be associated with new detections. In contrast, we show that through positively predicting and preserving trajectories of objects, the tracklets which lost their targets can be properly maintained for possible association in the future. 




%% file: 3_method.tex
\input{pipeline}
\section{Method}
Figure~\ref{fig:pipeline} shows the overview of our system. 
As an online tracking method, our method takes sequential detection results as input. 
For each frame, we predict the object locations for all tracklets with 3DKF.
Then we perform Hungarian algorithm for bipartite matching between detections and predictions. 
The bipartite matching process will output the matched pairs and the unmatched tracklets or detections.
The states of matched tracklets are updated by their corresponding detections, and unmatched tracklets are updated with their predicted object states.
The unmatched detections will be initialized as new tracklets. 
Each part in the pipeline will be described in detail in the following sections.

{\bf Detection.} 
By design, our \trackerspace is agnostic to the choice of detector.
For our best-reported result, We employ CenterPoint\cite{yin2021center} as our off-the-shelf detector.
CenterPoint chooses a quite permissive IoU threshold for NMS to ensure a better recall.
Therefore, we follow SimpleTrack~\cite{pang2021simpletrack} and perform a much stricter NMS on detection results to remove unwanted boxes before feeding them to our \tracker.

{\bf Trajectory Prediction.} 
We use a vanilla 3D Kalman filter(3DKF) for trajectory prediction. 
Following AB3DMOT\cite{Weng20203DMT}, we define the states of Kalman filter $Z=[x,y,z,\theta,l,w,h,\dot{x},\dot{y},\dot{z}]$ in 3D space, including object position, box size, box orientation and velocity.
Different from AB3DMOT, our object positions are defined in the world frame like \cite{yin2021center}, instead of in the LiDAR frame. 
The change of reference frame is crucial to our \tracker, helping us reduce the total number of tracklets managed.
During the tracking process, we alternate between predicting object state $X$ with 3DKF and updating 3DKF state $Z$ with incoming detection $D$. 
Both $X$ and $D$ are 7-dim vectors consisting of $[x, y, z, \theta, l, w, h]$. 
Note that we use $X$ to denote both a tracklet and the latest prediction of its associated 3DKF.
We may refer $X$ as tracklet or prediction interchangeable hereafter.

{\bf Detection-Prediction Association.}
For association, we compute 3D IoU or GIoU\cite{Rezatofighi_2019_CVPR, pang2021simpletrack} between detected 3D bounding boxes and boxes predicted by 3DKF. 
We then perform Hungarian matching based on 3D IoU / GIoU.
We discard low quality matchings if the 3D IoU / GIoU of a matching is lower than $\text{IoU}_\text{thres}$ or $\text{GIoU}_\text{thres}$. 
The inputs of the bipartite matching process is defined as:
\begin{align}
    \mathcal D&=\{D^1, D^2,\cdots, D^{p}\} \\
    \mathcal X&=\{X^1, X^2,\cdots, X^{q}\}
\end{align}
where $p$ and $q$ are the number of detections and tracklets in the frame.

The output of the bipartite matching process is defined as:
\begin{align}
    \mathcal D_\text{m}&=\{D_\text{m}^1, D_\text{m}^2, \cdots, D_\text{m}^{k}\}\\
    \mathcal D_\text{um}&=\{D_\text{um}^1, D_\text{um}^2, \cdots, D_\text{um}^{p-k}\}\\
    \mathcal X_\text{m}&=\{X_\text{m}^1, X_\text{m}^2, \cdots, X_\text{m}^{k_t}\}\\
    \mathcal X_\text{um}&=\{X_\text{um}^1, X_\text{um}^2, \cdots, X_\text{um}^{q-k}\}
\label{updatestep}
\end{align}

where $\mathcal X_\text{m}$ and $\mathcal D_\text{m}$ are the $k$ matched pairs of predictions and detections. 
$\mathcal X_\text{um}$ are the unmatched tracklets and $\mathcal D_\text{um}$ are the unmatched detections.

The matched detections $\mathcal D_\text{m}$ will be used to update the 3DKF states of corresponding tracklets $\mathcal X_\text{m}$. 
The unmatched tracklets $\mathcal X_\text{um}$ will use their own predictions to update 3DKF. 

{\bf Tracklet Birth and Preservation.} 
Following the common tracklet initialization strategy, we initialize a set of tracklets $\mathcal X_\text{new}$ for detections in $\mathcal D_\text{um}$.  
For these new tracklets, we will mark them as \emph{alive} if they are successfully associated with detections for $M_\text{hits}$ times in future frames. 
Otherwise, they stay at the \emph{birth} stage.

Different from previous works\cite{Weng20203DMT,weng2020gnn3dmot,sun2020scalability,chiu2020probabilistic,kim2020eager,yin2021center,benbarka2021score,zaech2021learnable} which terminate a tracklet after several frames since its last successful association, our \trackerspace always maintains tracklets for objects even if they are invisible.
The maintained tracklets will repeatedly predict their locations in the future frames with their last estimated states.
In this way, we maintain tracklets for all objects observed at least once. 
Our tracklets never die once created and hence are {\bf immortal}. 
And when the missing object reappears near its predicted trajectory, we can associate it with its corresponding tracklet again and update its tracklet with the observation instead of prediction. 

In each frame, we output tracklets in $\mathcal O=\{X_i \in \mathcal X_\text{m} \cap \mathcal X_\text{alive}\}$, where $\mathcal X_\text{alive}$ are tracklets in the alive state. 
Therefore the predictions of unmatched tracklets will not cause false positives. 
Algorithm ~\ref{fig:algorithm} summarizes our simple tracking algorithm.
\input{algorithm}

%% file: pipeline.tex
\begin{figure*}[t]
\begin{center}
\includegraphics[width=\linewidth,trim=0 10 0 0,clip]{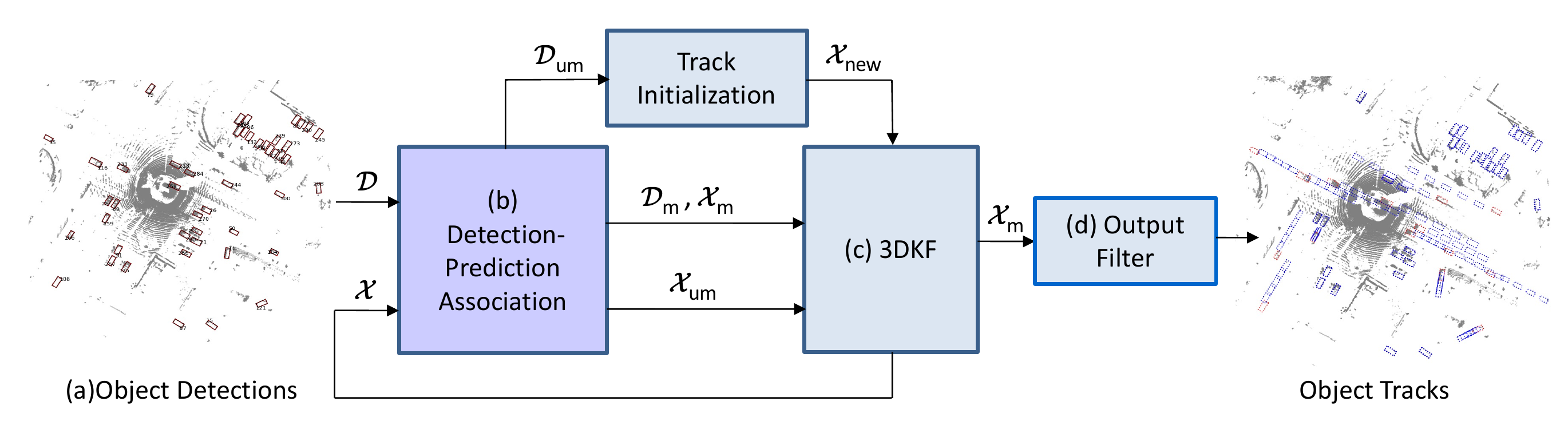}
\end{center}
\caption{
\textbf{Algorithm Pipeline.}
(a)We employ an off-the-shelf 3D detector to obtain 3D detections $\mathcal D$ from input point clouds.
(b)For data association, we compute 3D IoU or GIoU between detected 3D bounding boxes and boxes predicted by \tracker. Then we perform Hungarian matching based on 3D IoU/GIoU.
(c)We use a vanilla 3D Kalman filter(3DKF) for prediction. Based on the outputs of Hungarian matching process, The states of matched tracklets $\mathcal X_\text{m}$ are updated by its corresponding detection $\mathcal D_\text{m}$. The unmatched tracklets $\mathcal X_\text{um}$ will be updated with their predicted object states and the unmatched detections $\mathcal D_\text{um}$ will be initialized as new tracklets $\mathcal X_\text{new}$. 
(d)Only the alive tracklets which have left their birth stage and successfully been matched in the current frame are adopted as outputs.
}
\label{fig:pipeline}
\end{figure*}

%% file: algorithm.tex
\begin{algorithm}
\small
\caption{Tracking Pipeline}\label{alg:cap}
\renewcommand{\algorithmicrequire}{\textbf{Input:}}
\renewcommand{\algorithmicensure}{\textbf{Output:}}
\begin{algorithmic}
\Require 3D Object Detections $\mathcal{D}_{0:t}$
\Ensure Output Tracks $\mathcal{O}_{0:t}$ 
\State Frame number $t \gets 1$
\State Tracks set $\mathcal{X}_{0} \gets \{\}$
\While{not the end of tracking}
    \State ${\mathcal{X}}_t \gets \text{Prediction}(\mathcal{X}_{t-1})$
    \State $\mathcal{D}_\text{m},\mathcal{X}_\text{m},\mathcal{D}_\text{um},\mathcal{X}_\text{um} \gets \text{Match}(\mathcal{D}_t,\mathcal{X}_t) $
    
    \State $\mathcal{X}_\text{m} \xleftarrow{} \text{Update}({\mathcal D}_\text{m})$
    
    \State $\mathcal{X}_\text{new} \xleftarrow{} \text{Initialize}({\mathcal D}_\text{um})$
    
    \State $\mathcal O_t \gets \{X_i \in \mathcal X_\text{m} \cap \mathcal X_\text{alive}\}$
    \State $\mathcal{X}_{t+1} \gets \mathcal{X}_\text{m}\cup \mathcal{X}_\text{um} \cup \mathcal{X}_\text{new}$  
    \State $t \xleftarrow{} t + 1$
\EndWhile
\end{algorithmic}
\label{fig:algorithm}
\end{algorithm}


%% file: 4_results.tex

\section{Experiments}
\subsection{Datasets and Metrics.} 
We evaluate our method on Waymo Open Dataset(WOD) and nuScenes. 

{\bf Waymo Open Dataset} 
\cite{sun2020scalability} contains 1000 driving video sequences each lasts 20 seconds at 10 FPS. 
798 / 202 / 150 sequences are used for training, validation and testing, respectively. 
Point clouds and ground truth 3D boxes of objects in vehicle, pedestrian and cyclist classes are provided for each frame. 
Following the official evaluation metrics specified in \cite{sun2020scalability}, we report  Multiple Object Tracking Accuracy (MOTA)\cite{bernardin2008evaluating}, False Positives(FP), Miss and Mismatch for objects in the L2 difficulty. Readers may refer to \cite{bernardin2008evaluating} for detailed descriptions about these metrics.



{\bf nuScenes}\cite{caesar2020nuscenes} contains 1000 driving sequences in total with LiDAR scans and ground truth 3D box annotations provided at 20 FPS and 2 FPS, respectively. 
We report AMOTA\cite{weng2019baseline}, MOTA, and identity switches(IDS) for nuScenes. 
AMOTA is computed by integrating MOTA over different recalls, which is used as the primary metric for evaluating 3DMOT on nuScenes.

\subsection{Baselines}
{\bf CenterPoint}\cite{yin2021center} is both a 3D detection method and a 3DMOT method. Here we refer them as CenterPoint-DET and CenterPoint-3DMOT, respectively. 
We compare our proposed method with CenterPoint-3DMOT since we use the same CenterPoint-DET detection results.
Also, CenterPoint-3DMOT serves as a very strong baseline on both WOD and nuScenes.

{\bf CenterPoint++} is our modified tracking implementation of CenterPoint. CenterPoint-3DMOT only performs constant velocity trajectory prediction and nearest box center matching. It is unfair to compare it with our method which uses 3DKF for trajectory prediction and 3D IoU / GIoU for matching. 
Therefore we set another baseline named CenterPoint++, which uses 3DKF and 3D IoU / GIoU matching like ours. The main difference between CenterPoint++ and \trackerspace is that a traditional tracklet termination mechanism is used in CenterPoint++. We terminate a tracklet in CenterPoint++ after $A_\text{max}$ frames since its last successful association following previous works\cite{Weng20203DMT,weng2020gnn3dmot,chiu2020probabilistic,kim2020eager,yin2021center,zaech2021learnable}. All the other hyper-parameters of CenterPoint++ are set the same as our \trackerspace for a fair comparison.

\subsection{ Implementation Details}
We take CenterPoint detection boxes as input. 
We drop boxes with scores below 0.5 on WOD.
We keep all boxes on nuScenes.
We also perform 3D IoU-based NMS on detected bounding boxes with 0.25 / 0.1 3D IoU thresholds on WOD / nuScences, respectively. 
In detection-prediction association, we use 3D IoU / GIoU between detections and predictions on WOD/nuScenes, respectively. 
Then we perform Hungarian algorithm over the 3D IoU distance matrix. 
We drop all detection-prediction associations with 3D IoU below 0.1 or 3D GIoU below -0.5. 
For tracklet initialization, we set the hits to birth $M_\text{hits}=1$ on both WOD and nuScenes. 
For CenterPoint++, we set $A_\text{max}=2$.

\subsection{Main Results}
We compare \trackerspace to previous 3DMOT methods on both WOD and nuScenes.
\input{waymotest}
\input{waymo_val}
\input{nuscenes_test}

\input{nuscenes_val}
{\bf Results on WOD.} 
Table~\ref{tab:waymo} and ~\ref{tab:waymo_val} show comparisons of tracking results between our method, our baselines and other state-of-the-art methods on the WOD test and validation sets. 
For the vehicle class, our method reduces the mismatch ratio to a new level of 0.0001, several times lower than previous works. 
It is noteworthy that our method performs much better than CenterPoint++ in the mismatch metric. 
This indicates the majority of mismatch cases we have reduced are results of maintaining the tracklets with predicted trajectories rather than employing 3DKF as motion model.
Besides the substantial improvement our proposed method achieved on the mismatch ratio, we also achieve a 0.4/0.9 MOTA improvement over CenterPoint++ for vehicle/pedestrian, respectively, outperforming all previously published methods.


{\bf Results on nuScenes.} 
Table~\ref{tab:nuscenes} and ~\ref{tab:nuscenes_val} shows our results on nuScenes test and validation sets. 
We compare our \trackerspace with published LiDAR-only works for a fair comparison. 
All methods in Table~\ref{tab:nuscenes} and ~\ref{tab:nuscenes_val} are based on the same CenterPoint detection results, allowing a fair comparison between different tracking approaches. 
Our method achieves a 2.3 AMOTA improvement over CenterPoint and a 0.3 improvement over CenterPoint++. 

\subsection{Ablation Studies}
\input{ablation_detector}
We analyze design choices and hyper-parameter settings of \trackerspace on the WOD validation set for the vehicle class. 
Results show that our \trackerspace is robust to a wide range of design choices and hyper-parameters. 

{\bf The performance of base 3D detectors.} Table~\ref{tab:detector} summarizes the performances of our proposed method on top of different 3D detectors on the WOD validation set. 
We take detection results of several 3D detectors including PointPillars+PPBA\cite{lang2019pointpillars,cheng2020improving}
, SECOND\cite{yan2018}, LiDAR R-CNN\cite{li2021lidar} and CenterPoint that take voxelized or Bird’s Eye View (BEV) representations of point clouds for object detection, and RangeDet\cite{fan2021rangedet} which takes range view representation of collected 3D informations as input. 
The results show that our method is agnostic to 3D detectors and can reduce the mismatch rates to 0.0001 level on top of various detectors. 

{\bf When to terminate a tracklet. } 
\input{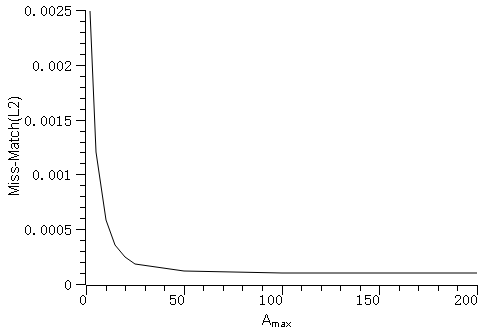}
\trackerspace permanently maintains tracklets for objects gone dark, which is the main difference between \trackerspace and CenterPoint++. For ablation experiment, here we introduce the trajectory prediction as well as tracklets preservation mechanisms into CenterPoint++, but still with a max preservation age $A_\text{max}$ for the tracklets. Then we increase $A_\text{max}$ from 2 to study the influence of premature tracklet termination.

As shown in Figure~\ref{fig:age}, the mismatch ratio drop significantly and monotonically when we extend the life of tracklets by increasing $A_\text{max}$. 
When we preserve tracklets with predicted trajectories for 50 frames the mismatch ratio reaches 0.000123, 20 times lower than when the tracklets are preserved by predictions for only 2 frames. 
And when the life of tracklet continues to grow from 50 frames, up to 17\% of remaining mismatch cases can further be avoided. 
This indicates that preserving tracklets by trajectory prediction can significantly reduce mismatch cases without introducing significant tracklets confusions.


{{\bf The minimum birth count}  $M_{\text{hits}}$.} 
\input{abl_birth}
We also explore the influence of the minimum birth count $M_\text{hits}$ on the performance of our proposed method. The results are shown in Table~\ref{tab:birth}. Clearly \trackerspace is robust to $M_\text{hits}$.

{\bf Association Threshold.}
We find the association threshold $\rm IoU_{thres}$ is a major influence factor to the performance of \tracker. As shown in Table~\ref{tab:assothres},  when we increase $\rm IoU_{thres}$, \trackerspace reports significantly more mismatch cases with a lower MOTA. 
\input{abl_assothres}

{\bf Redundancy of input detections.}
In our method, we perform NMS on detection results with a relatively low IoU threshold compared to CenterPoint. 
The motivation is that considering rigid objects will not overlap in 3D space, the overlapped 3D detections have a high probability to be false positives and shall be discarded. 
Our experiment results in Table~\ref{tab:nmsthres} show when we decrease the NMS threshold for a more strict suppression, the mismatch ratio drops by a large margin.
This result supports our assumption. 
	
\input{abl_nmsthres}

\subsection{Qualitative Results.}
Fig~\ref{failcase} shows the BEV visualization of the only remaining wrong association case in Immortal Tracker. 
Fig~\ref{successcase} visualizes one of the prevented premature terminations in Immortal Tracker.
\input{failcase}

\input{successcase}

%% file: waymotest.tex
\begin{table*}\normalsize
\centering
\resizebox{\linewidth}{17mm}{ 
\begin{tabular}{@{}lcccc|cccc@{}}
\toprule
 \multirow{2}*{Method}& \multicolumn{4}{c}{Vehicle}& \multicolumn{4}{c}{Pedestrian}  \\
 
  & MOTA\%$\uparrow$ & FP\%$\downarrow$ & Miss\%$\downarrow$  & Mismatch\%$\downarrow$  & MOTA\%$\uparrow$ & FP\%$\downarrow$ & Miss\% $\downarrow$  & Mismatch\%$\downarrow$  \\
\midrule
AB3DMOT\cite{sun2020scalability} & 40.1 & 16.4 & 43.4 & 0.13 & 37.7 & 11.6 & 50.2 & 0.47\\
CenterPoint\cite{yin2021center} &59.4&9.4&\textbf{30.9}&0.32&56.6&\textbf{9.3}&33.1&1.07 \\
SimpleTrack*\cite{pang2021simpletrack} & 60.3 & 8.8 & \textbf{30.9} & 0.08 & 60.1 & 10.7 & 28.8 & 0.40\\
CenterPoint++ &60.2&\textbf{8.4}&31.2&0.24&59.7&10.3&29.2&0.85\\
\tracker(Ours) &\textbf{60.6}&8.5&31.0&\textbf{0.01}&\textbf{60.6}&11.0&\textbf{28.3}&\textbf{0.18}\\
\bottomrule
\end{tabular}
} 
\caption{
State-of-the-art comparisons for 3D MOT on WOD test set. * represents for works that are not peer-reviewed yet
} 
\label{tab:waymo}
\end{table*}

%% file: waymo_val.tex
\begin{table*}\normalsize
\centering
\resizebox{\linewidth}{!}{ 
\begin{tabular}{@{}lcccc|cccc@{}}
\toprule
 \multirow{2}*{Method}& \multicolumn{4}{c}{Vehicle}& \multicolumn{4}{c}{Pedestrian}  \\
 
  & MOTA\%$\uparrow$ & FP\%$\downarrow$ & Miss\%$\downarrow$  & Mismatch\%$\downarrow$  & MOTA\%$\uparrow$ & FP\%$\downarrow$ & Miss\% $\downarrow$  & Mismatch\%$\downarrow$  \\
\midrule

CenterPoint\cite{yin2021center} &55.1 & 10.8 & 33.9 & 0.26 & 54.9 & \textbf{10.0} & 34.0 & 1.13\\
SimpleTrack*\cite{pang2021simpletrack} & 56.1 & 10.4 & \textbf{33.4} & 0.08 & 57.8 & 10.9 & 30.9 & 0.42\\
CenterPoint++ &56.1 & \textbf{10.2} & 33.5 & 0.25& 57.4 & 11.1 & 30.6 & 0.94\\
\tracker(Ours) &\textbf{56.4} & \textbf{10.2} & \textbf{33.4} & \textbf{0.01}&\textbf{58.2}&11.3&\textbf{30.5}&\textbf{0.26}\\
\bottomrule
\end{tabular}
} 
\caption{
State-of-the-art comparisons for 3D MOT on WOD validation set. * represents for works that are not peer-reviewed yet
} 
\label{tab:waymo_val}
\end{table*}

%% file: nuscenes_test.tex
\begin{table}\normalsize
\centering
\resizebox{\linewidth}{!}{ 
\begin{tabular}{@{}lcccccc@{}}
\toprule

  Method & AMOTA\%$\uparrow$ & MOTA\%$\uparrow$ &  IDS$\downarrow$ \\
\midrule
StanfordIPRL-TRI\cite{chiu2020probabilistic} & 55.0 & 45.9 & 950\\
CenterPoint\cite{yin2021center} & 63.8 & 53.7  & 760 \\
CBMOT\cite{benbarka2021score} & 64.9 & 54.4 & 557\\ 
OGR3MOT*\cite{zaech2021learnable}  & 65.6 & 55.4 &  \textbf{288} \\
CenterPoint++  & 65.8 & \textbf{55.7} & 609\\
\tracker(Ours)        & \textbf{66.1} & 55.2  & 365\\
\bottomrule
\end{tabular}
} 
\caption{
State-of-the-art comparisons for 3D MOT on nuScenes test set. Only published online methods which are Lidar-based are reported. * represents for works that are not peer-reviewed yet.
} 
\label{tab:nuscenes}
\end{table}

%% file: nuscenes_val.tex
\begin{table}\normalsize
\centering
\resizebox{\linewidth}{15mm}{ 
\begin{tabular}{@{}lcccccc@{}}
\toprule

  Method & AMOTA\%$\uparrow$ & MOTA\%$\uparrow$ &  IDS$\downarrow$ \\
\midrule
CenterPoint\cite{yin2021center} & 66.5 & 56.2  & 562 \\
CBMOT\cite{benbarka2021score} & 67.5 & 58.3 & 494\\ 
OGR3MOT*\cite{zaech2021learnable}  & 69.3 & \textbf{60.2} &  \textbf{262} \\
CenterPoint++  & 68.7 & 59.2 & 519\\
\tracker(Ours)        & \textbf{70.2} & 60.1  & 385\\
\bottomrule
\end{tabular}
} 
\caption{
State-of-the-art comparisons for 3D MOT on nuScenes validation set. Only published online methods which are Lidar-based are reported. * represents for works that are not peer-reviewed yet.
} 
\label{tab:nuscenes_val}
\end{table}

%% file: ablation_detector.tex
\begin{table*}\normalsize
\centering
\begin{tabular}{@{}lccccc@{}}
\toprule
 Detector & MOTA\%$\uparrow$ & FP(\%)$\downarrow$ & Miss(\%)$\downarrow$  & Mismatch(\%)$\downarrow$  \\
\midrule
PointPillars+PPBA\cite{lang2019pointpillars,cheng2020improving}++ & 40.9 & 9.3 & 49.5 & 0.34\\
PointPillars+PPBA + \tracker & 41.3 & 9.4 & 49.3 & 0.01\\
\midrule
SECOND\cite{yan2018}++ & 50.2 & 8.6 & 40.9 & 0.31\\
SECOND + \tracker & 50.6 & 8.7 & 40.8 & 0.01\\
\midrule
RangeDet\cite{fan2021rangedet}++ & 50.5 & 10.2 & 39.1 & 0.25\\
RangeDet + \tracker & 50.7 & 10.2 & 39.1 & 0.02 \\
\midrule
Lidar R-CNN\cite{li2021lidar}++ & 53.7 & 7.3 & 38.1 & 0.84\\
Lidar R-CNN + \tracker & 54.9 & 8.2 & 36.9 & 0.02 \\
\midrule
CenterPoint++& 56.1 & 10.2 & 33.5 & 0.25\\
CenterPoint + \tracker& 56.4 & 10.2 & 33.4 & 0.01\\
\bottomrule
\end{tabular}
\caption{
Ablation study for different detectors on WOD validation set. Here ++ represents for we apply the same baseline tracker described in CenterPoint++ on top of the detector.
} 
\label{tab:detector}
\end{table*}

%% file: age.tex
\begin{figure}[t]
\begin{center}
\includegraphics[width=\linewidth]{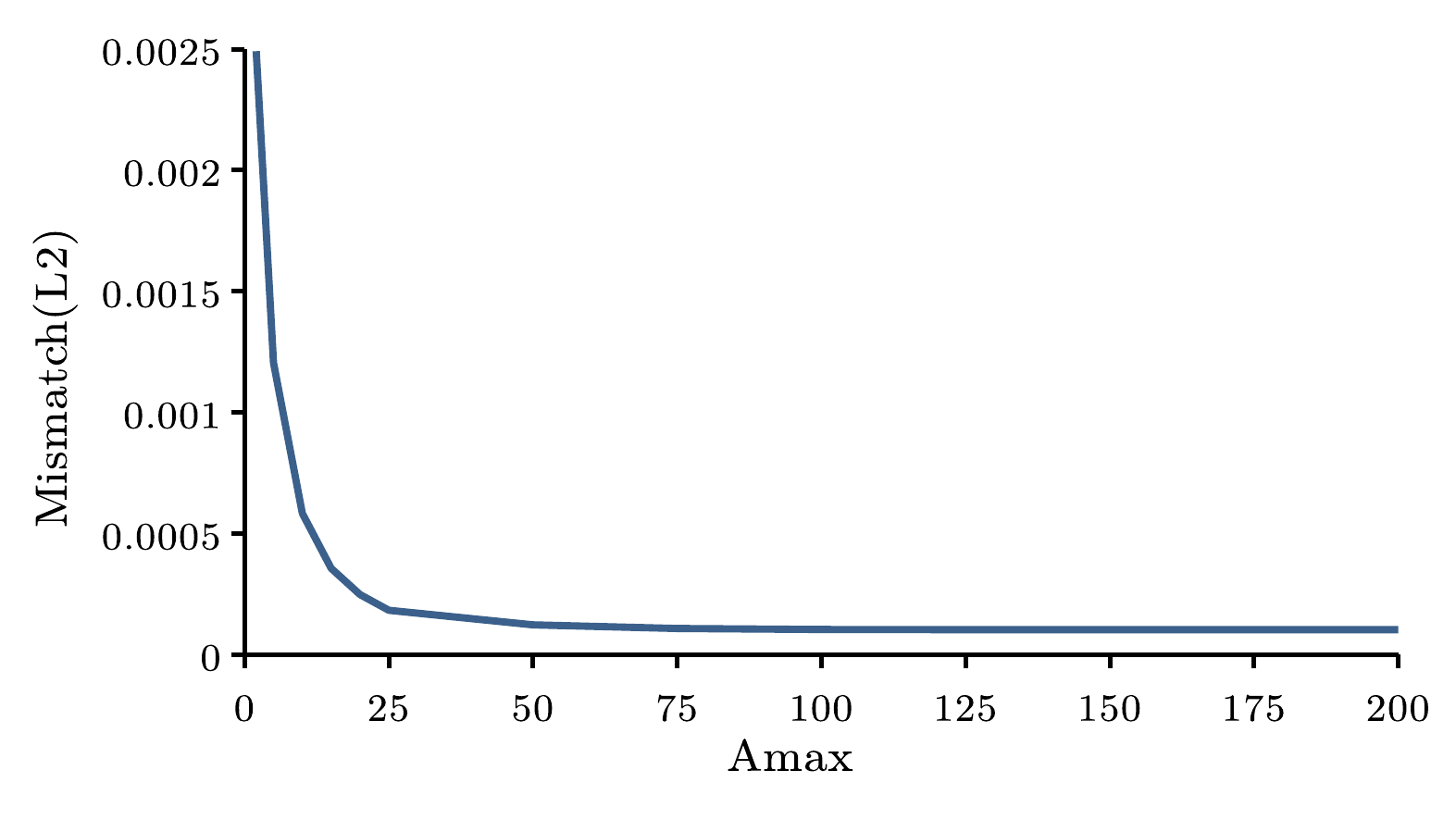}
\end{center}
\caption{
Details for mismatch ratio vs $A_\text{max}$ on WOD validation set.
}
\label{fig:age}
\end{figure}

%% file: abl_birth.tex
\begin{table}\normalsize
\centering
\resizebox{\linewidth}{!}{ 
\begin{tabular}{@{}lcccc@{}}
\toprule
 $M_{\text{hits}}$  & MOTA\%$\uparrow$ & FP\%$\downarrow$ & Miss\%$\downarrow$  & Mismatch\%$\downarrow$  \\
\midrule
 0 & 56.4 & 10.3 & 33.4 & 0.01\\
 1 & 56.4 & 10.2 & 33,4 & 0.01\\
 2 & 56.4 & 10.2 & 33.4 & 0.01\\
 3 & 56.3 & 10.2 & 33.5 & 0.01\\
\bottomrule
\end{tabular}
} 
\caption{
Ablation study for the minimum birth count $M_{\text{hits}}$ on WOD validation set.
} 
\label{tab:birth}
\end{table}

%% file: abl_assothres.tex
\begin{table}\normalsize
\centering
\resizebox{\linewidth}{!}{ 
\begin{tabular}{@{}lcccc@{}}
\toprule
 {$\rm IoU_{thres}$}  & MOTA\%$\uparrow$ & FP\%$\downarrow$ & Miss\%$\downarrow$  & Mismatch\%$\downarrow$  \\
\midrule
 0.7 & 51.5 & \textbf{9.6} & 38.1 & 0.86\\
 0.5 & 54.7 & 10.0 & 34.9 & 0.39\\
 0.3 & 56.2 & 10.2 & 33.5 & 0.08\\
 0.1 & \textbf{56.4} & 10.2 & \textbf{33.4} & \textbf{0.01}\\
\bottomrule
\end{tabular}
} 
\caption{
Ablation study for the association threshold $\rm IoU_{thres}$ on WOD validation set.
} 
\label{tab:assothres}
\end{table}

%% file: abl_nmsthres.tex
\begin{table}\normalsize
\centering
\resizebox{\linewidth}{!}{ 
\begin{tabular}{@{}lcccc@{}}
\toprule
 $\rm NMS_{thres}$  & MOTA\%$\uparrow$ & FP\%$\downarrow$ & Miss\%$\downarrow$  & Mismatch\%$\downarrow$  \\
\midrule
 0 & 56.3 & 10.2 & 33.5 & \textbf{0.01}\\
 0.25 & {56.4} & 10.2 & {33.4} & \textbf{0.01}\\
 0.5 & 56.4 & 10.2 & {33.4} & 0.02\\
 0.75 & 56.2 & 10.2 & {33.4} & 0.15\\
\bottomrule
\end{tabular}
} 
\caption{
Ablation study for the NMS threshold $\rm NMS_{thres}$ on WOD validation set.
} 
\label{tab:nmsthres}
\end{table}

%% file: failcase.tex
\begin{figure*}
	\subfloat[frame 108]{
		\includegraphics[scale=0.13]{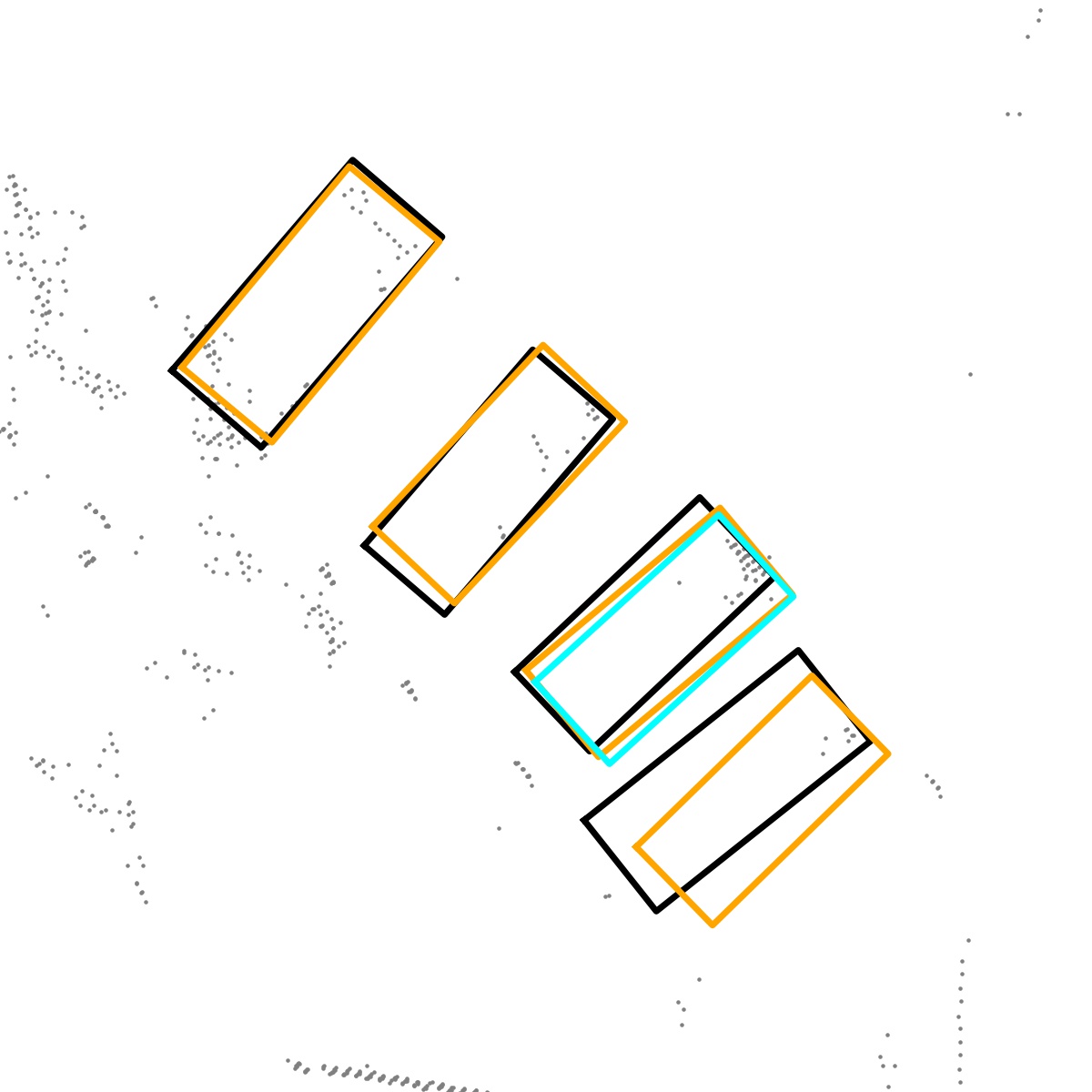}}
	\subfloat[frame 112]{
		\includegraphics[scale=0.13]{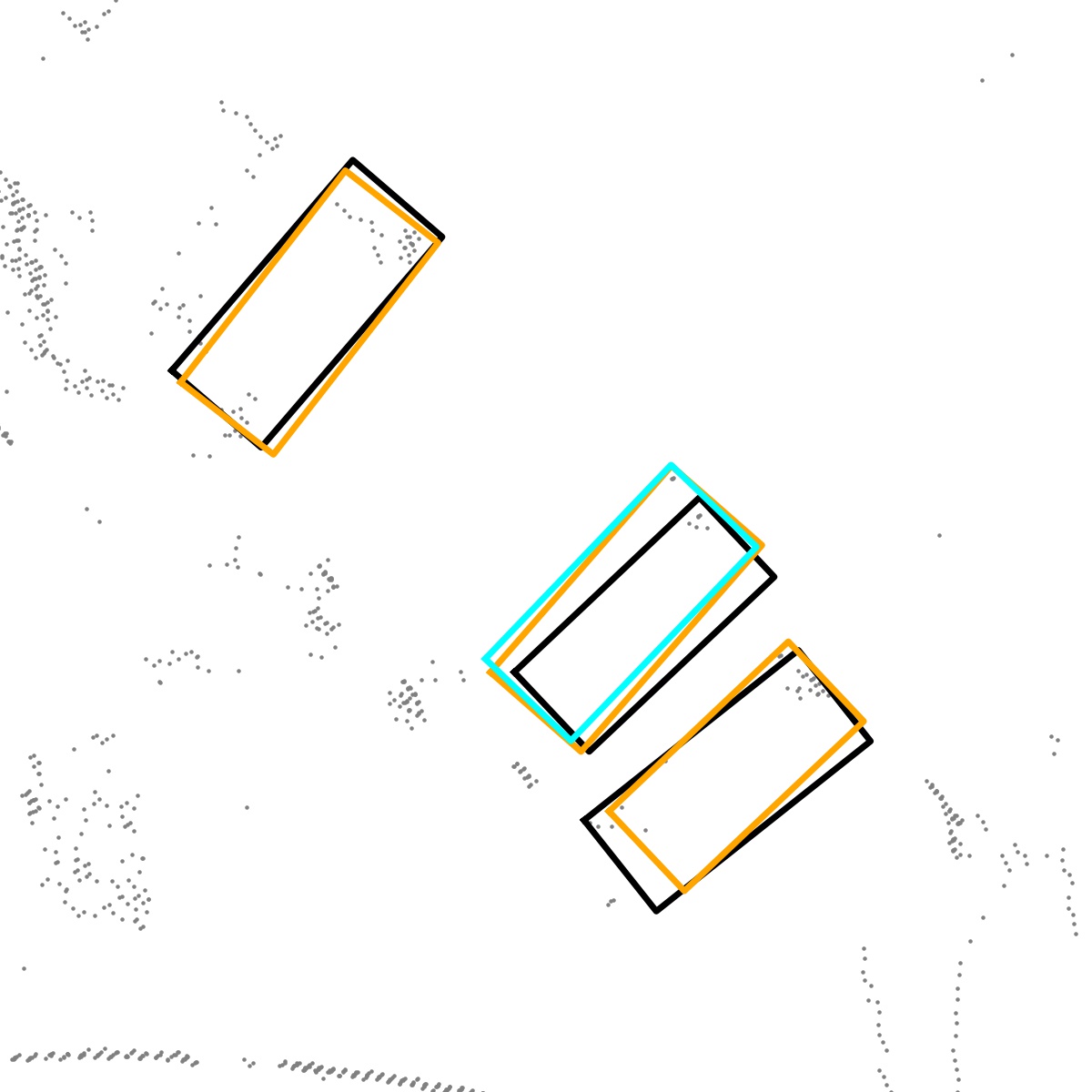}}
	\subfloat[frame 114]{
		\includegraphics[scale=0.13]{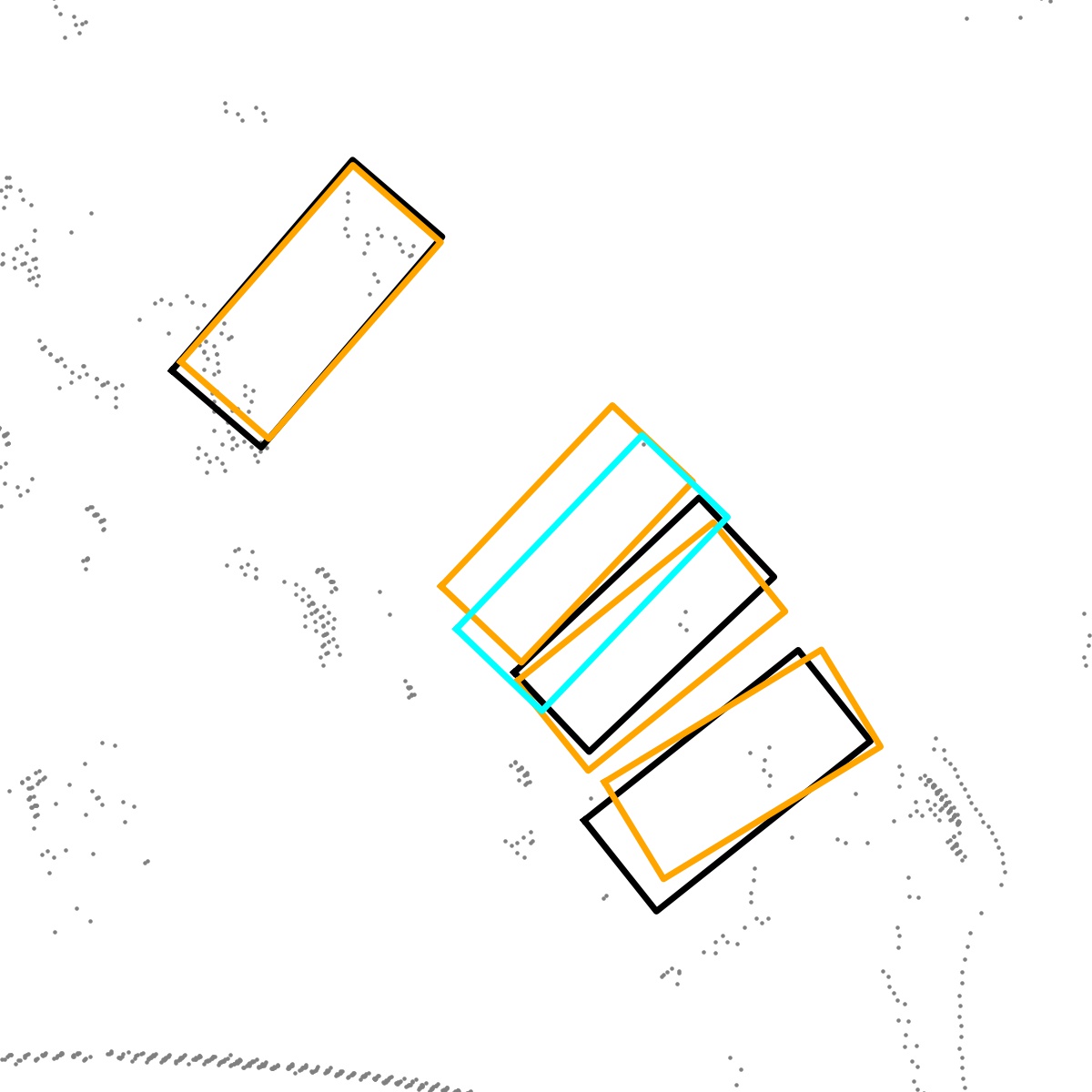} }
	\subfloat[frame 118]{
		\includegraphics[scale=0.15]{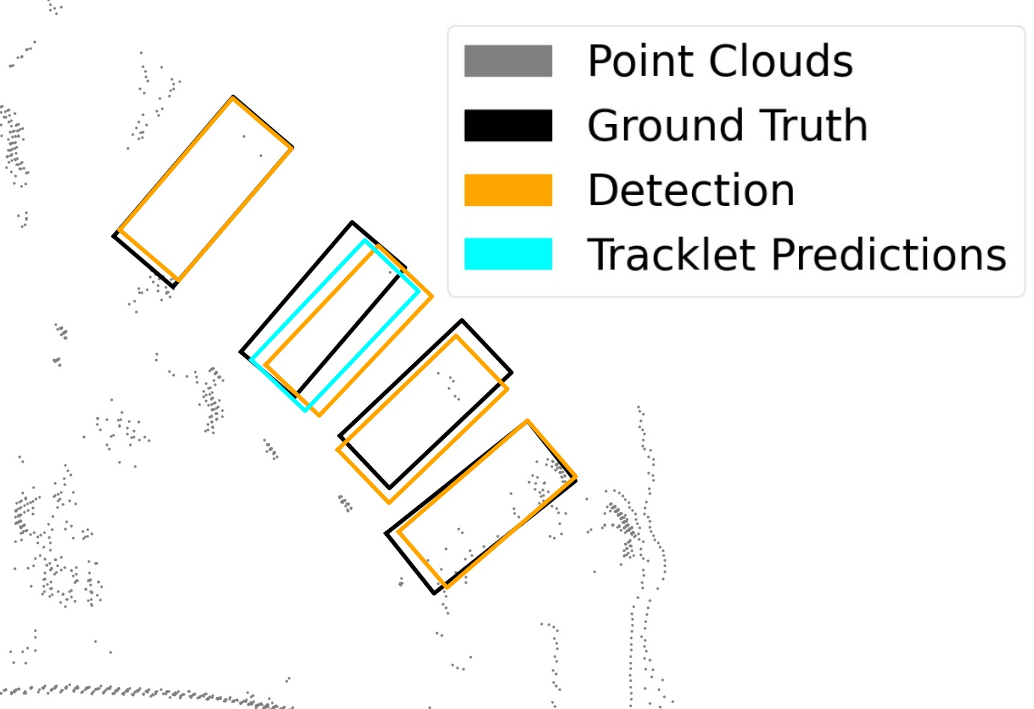}}
	\caption{Visualization of the remaining wrong association case. Outputs of the wrongly associated tracklet are drawn in cyan. The ground truth bounding boxes are drawn in black and the detection results are drawn in brown. The tracklet is misled by sequential inaccurate or false positive detections from its initial location to another. Such a case is extremely rare in our method.}
	\label{failcase} 
\end{figure*}

%% file: successcase.tex
\begin{figure*}
\begin{center}
	\subfloat[CenterPoint++]{
		\includegraphics[scale=0.22]{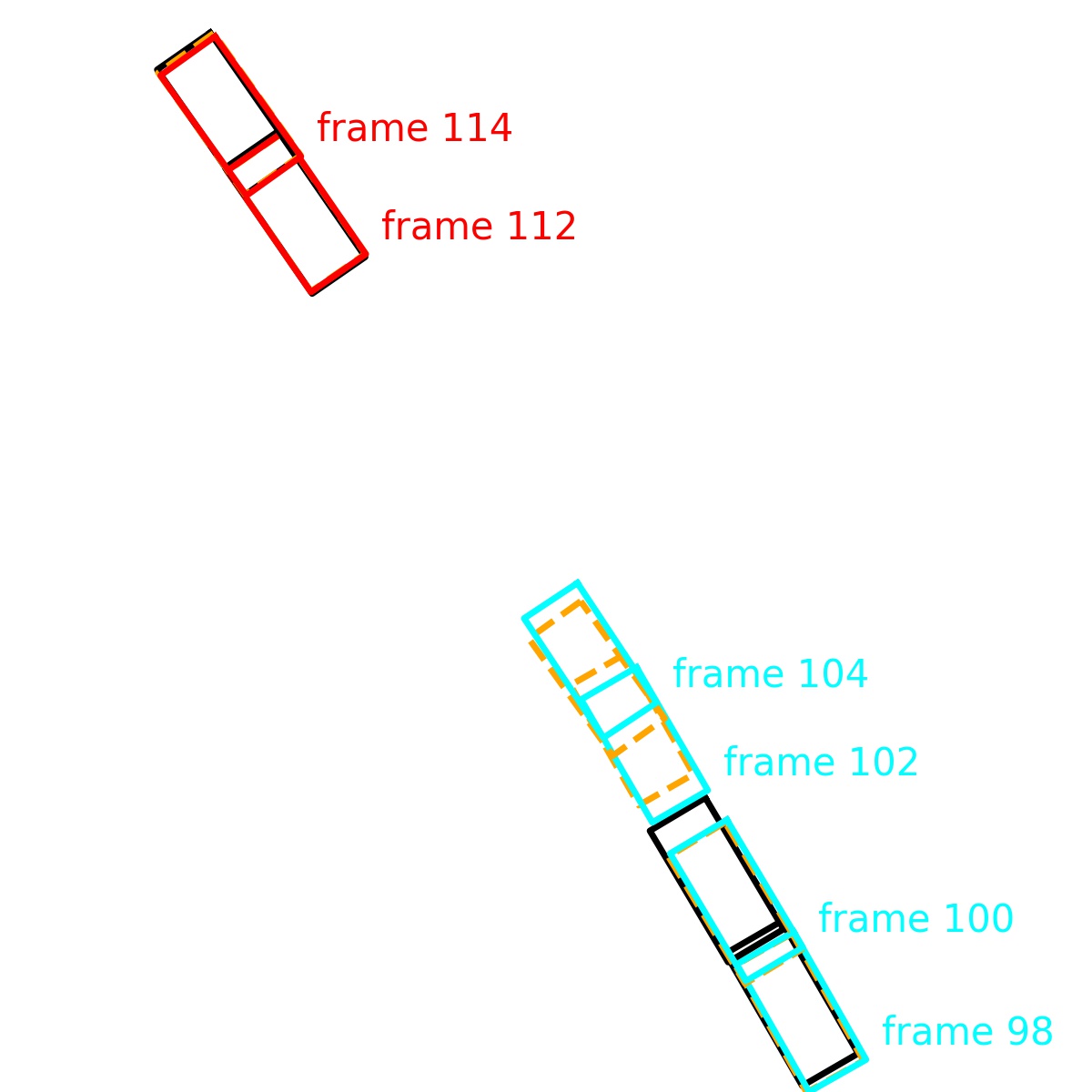}}
	\subfloat[Immortal Tracker]{
		\includegraphics[scale=0.22]{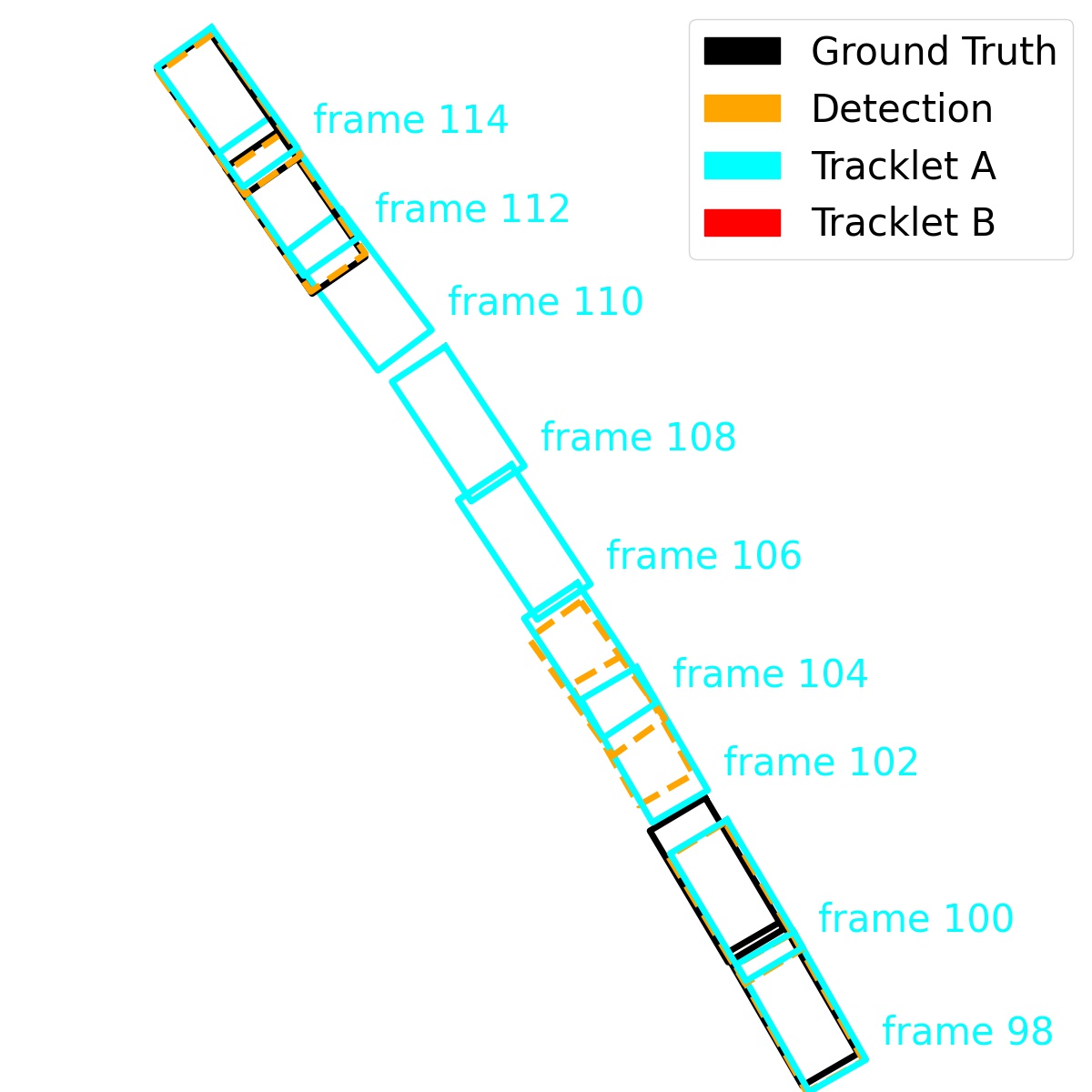}
		}

	\caption{Visualization of a prevented premature termination. In this case the vehicle is not detected in frame 100-112. We plot the estimated bounding boxes of tracklets every two frames. Tracklets with different identities are drawn in different colors. For a better view we do not plot the ground truth bounding boxes in frame 100-112. (a)In CenterPoint++, the tracklet(marked in cyan) is terminated when the object is temporarily not observed, causing an identity switch. (b)While in Immortal Tracker, the initial tracklet is preserved until the object is detected again.
	}
	\label{successcase} 
\end{center}
\end{figure*}

%% file: 5_conclusions.tex
\section{Conclusion and Future Direction}
In this work, we found that identity switches in 3D MOT are almost solely caused by premature tracklet termination widely existing in modern 3D MOT methods. 
We proposed using trajectory prediction to preserve tracklets for invisible objects.
We found using a simple 3D Kalman filter for trajectory prediction could reduce identity switch cases by 96\%.
Our method provided valuable insights for handling long-standing challenges in tracking like long-term occlusion.
However, due to the limitation of existing 3DMOT benchmarks, we can not verify the effectiveness of our predict-to-track paradigm in long-term($\gg 20$ seconds) and/or more interactive(\textit{e.g.} the famous Shibuya crossing) scenarios. 
In the future, we want to set up a more challenging benchmark for 3DMOT and explore the synergy between trackers and more sophisticated trajectory predictors.

